\documentclass[twoside,twocolumn]{article}  
\usepackage{ltexpprt} 
\usepackage[utf8]{inputenc} 
\usepackage{array} 
\usepackage{graphicx,color,enumerate} 
\usepackage{hyperref}
\usepackage{graphicx}
\usepackage{wrapfig,lipsum,booktabs,caption}

\usepackage{multirow}
\usepackage{amsmath,amsfonts,amssymb}
\usepackage{algpseudocode}
\usepackage{varwidth}% http://ctan.org/pkg/varwidth
\usepackage{algorithm}

\usepackage{epstopdf}

\newcommand{\mb}[1]{\mbox{\boldmath $#1$}}

\newcommand{\bx}{\mb{x}}

\newcommand{\OO}{\mathcal{O}}
\newcommand{\CC}{\bf{C}}
\newcommand{\JJ}{\mathcal{J}}

\usepackage{xspace}

\usepackage[inline]{trackchanges}
\addeditor{tr}

\begin{document}

\title{\Large Fast Multilevel Support Vector Machines}%\thanks{Supported by GSF grants ABC123, DEF456, and GHI789.}}
\author{Talayeh Razzaghi\thanks{trazzag@clemson.edu, School of Computing, Clemson University, Clemson, SC} \\
\and 
Ilya Safro\thanks{isafro@clemson.edu, School of Computing, Clemson University, Clemson, SC}}
\date{}

\maketitle

%\pagenumbering{arabic}
%\setcounter{page}{1}%Leave this line commented out.

\begin{abstract} \small\baselineskip=9pt 
%Papers must have an abstract with a maximum of 300 words and a keyword list with no more than 6 keywords
Solving different types of optimization models (including parameters fitting) for support vector machines on large-scale training data is often an expensive computational task. This paper proposes a multilevel algorithmic framework that scales efficiently to very large data sets. Instead of solving the whole training set in one optimization process, the support vectors are obtained and gradually refined at multiple levels of coarseness of the data. The proposed framework includes: (a) construction of hierarchy of large-scale data coarse representations, and  (b) a local processing of updating the hyperplane throughout this hierarchy. Our multilevel framework substantially improves the computational time without loosing the quality of classifiers. The algorithms are demonstrated for both regular and weighted support vector machines. Experimental results are presented for balanced and imbalanced classification problems. Quality improvement on several imbalanced data sets has been observed.


\end{abstract}

\section{Introduction}

Support vector machines (SVM) are among the most well-known optimization-based supervised learning methods, originally developed for binary classification problems \cite{vapnik2000nature}. The main idea of SVM is to identify a decision boundary with maximum possible margin between the data points of each class. 
%There are certain parameters that needed to be tuned using model selection techniques.
Training nonlinear SVMs is often a time consuming task when the data is large. This problem becomes extremely sensitive when the model selection techniques are applied.
%and model selection become a time consuming process while implemented to very large datasets. 
Requirements of computational resources, and storage are growing rapidly with the number of data points, and the dimensionality, making many practical classification problems less tractable. In practice, when solving SVM, there are several parameters that have to be tuned. Advanced methods, such as the grid search and the uniform design for tuning the parameters, are usually implemented using iterative techniques, and the total complexity of the SVM strongly depends on these methods, and on the quality of the employed optimization solvers \cite{chang2011libsvm}. 

In this paper, we focus on SVMs that are formulated as the convex quadratic programming (QP). Usually, the complexity required to solve such SVMs is between $\OO(n^2)$ to $\OO(n^3)$ \cite{graf2004parallel}. For example, the popular QP solver implemented in LibSVM \cite{chang2011libsvm} scales between $\OO(n_{f}{n_{s}}^2)$ to $\OO(n_{f}{n_{s}}^3)$ subject to how efficiently the LibSVM cache is employed in practice, where $n_f$, and $n_s$ are the numbers of features, and samples, respectively. 
Typically, gradient descent methods achieve good performance results, but still tend to be
very slow for large-scale data (when effective parallelism is hard to achieve). Several works have recently addressed this problem. Parallelization usually splits the large set into smaller subsets and then performs a training to assign data points into different subsets \cite{collobert2002parallel}. In \cite{graf2004parallel}, a parallel version of the Sequential Minimal Optimization (SMO) was developed to accelerate the solution of QP. Although parallelizations over the full data sets often gain good performance, they can be problematic to implement due to the dependencies between variables, which increases communication. Moreover, although specific types of SVMs might be appropriate for parallelization (such as the Proximal SVM \cite{tveit2003multicategory}), the question of their practical applicability for high-dimensional datasets still requires further investigation.
Another approach to accelerate the QP is chunking \cite{Joachims1999,catak2013cloudsvm}, in which the optimization problem is solved iteratively on the subsets of training data until the global optimum is achieved.
The SMO is among the most popular methods of this type \cite{platt1999fast}, which scales down the chunk size to two vectors. Shrinking to identify the non-support vectors early, during the optimization, is another common method that significantly reduces the computational time \cite{Joachims1999,chang2011libsvm,collobert2002torch}. Such techniques can save substantial amounts of storage when combined with caching of the kernel data.  
%\note[tl]{I deleted the digesting because it is more related to parallel methods for training SVM, and I am concerned if the reviewers might ask to compare our method with these methods that we talked here in the introduction? \\
Digesting is another successful strategy that "optimizes subsets of training data closer to completion before adding new data" \cite{decoste2002training}.

Imbalanced classification problems (when the sizes of classes are very different) are another major problem that, in practice, can lead to poor performance measures \cite{tang2009svms}.  Imbalanced learning is a significant emerging problem in many areas, including medical diagnosis  \cite{lo2008learning,mazurowski2008training,li2010learning}, face recognition \cite{kwak2008feature}, bioinformatics \cite{batuwita2009micropred}, risk management \cite{ezawa1996learning,groth2011intraday}, and manufacturing \cite{su2007evaluation}. Many standard SVM algorithms often tend to misclassify the data points of the minority class. One of the most well-known techniques to deal with imbalanced data is the cost-sensitive learning (CSL). The CSL addresses imbalanced classification problems through different cost matrices. The adaptation of cost-sensitive learning with the regular SVM is known as \emph{weighted support vector machine} (WSVM, also termed as Fuzzy SVM) \cite{lin2002fuzzy}.

%These type of sets suppose a new challenging problem for Data Mining, since standard classification algorithms usually consider a balanced training set and this supposes a bias towards the majority class.

In this paper, we propose a novel method for efficient and effective solution of SVM and WSVM. In the heart of this method lies a multilevel algorithmic framework (MF) inspired by the multiscale optimization strategies \cite{brandt:optstrat}.
The main objective of multilevel algorithms is to construct a hierarchy of problems (coarsening), each approximating the original problem but with fewer degrees of freedom. This is achieved by introducing a chain of successive projections of the problem domain into lower-dimensional or smaller-size domains and solving the problem in them using local processing (uncoarsening). The MF combines solutions achieved by the local processing at different levels of coarseness into one global solution. Such frameworks have several key advantages that make them  attractive for applying on large-scale data: it exhibits a linear complexity,
and it can be relatively easily parallelized. Another advantage
of the MF is its heterogeneity, expressed in the ability to incorporate external appropriate optimization algorithms (as a refinement) in the framework at different levels. These frameworks were extremely successful in various practical machine learning and data mining tasks such as clustering \cite{mlmodul,KushnirGB06}, segmentation \cite{sharon06Hierarchy}, and dimensionality reduction \cite{raey}.

The contribution of this paper is a novel multilevel algorithmic approach for developing scalable SVM and WSVM. We propose a multilevel algorithm that creates a hierarchy of coarse representations of the original large-scale training set, solves SVM (or WSVM) at the coarsest level where the number of data points is small, and prolongates the hyperplane throughout the created hierarchy by gradual refinement of the support vectors. The proposed method admits an easy  parallelization, and its superiority is demonstrated through extensive computational experiments. The method requires considerably less memory than regular SVMs. The method is particularly successful on imbalanced data sets as it creates a balanced coarse representation of the problem that allows to effectively approximate the separating hyperplane for the original problem.

%The remainder of the paper is organized as follows. In Sect. 2, we first review the standard
%SVM and WSVM formulations, and then, we present the hierachial multilevel framework for SVM and WSVM with three main phases.
%In Sect. 3, we present Multilevel SVM and  WSVM results for binary classification, and present results in comparison to the
%regular single-level methods. The conclusion is presented in Sect. 4.

% the author found that the most time-consuming part is located in the kernel evaluation. Therefore, he parallelized the kernel evaluations and gradient updates, combined with inner sequential QP solver and distributed storage of kernel rows, achieved linear speedup for regression in his test.

\section{Problem Definition}

%In this section, we present the multilevel framework of SVM and WSVM.

%SVM and WSVM formulation, which combines the ideas behind the multilevel formulation to reduce the size of training data for training and parameter tuning purposes and the robustness of SVM-based formulation for both balanced and imbalanced classification problems. 

%\paragraph{Support Vector Machines}
Let a set of labeled data points be represented by a set  $\mathcal{J}=\{(x_i, y_i)\}_{i=1}^l$, where $(x_i, y_i)~\in~\mathbb{R}^{n+1}$, and  $l$ and $n$ are the numbers of data points and features, respectively. Each $x_i$ is a data point with $n$ features, and a class label $y_i \in \{-1,1\}$. Subsets of $\JJ$ related to the ``majority'' and ``minority'' classes are denoted by $\CC^+$, and $\CC^-$, respectively, i.e., $\mathcal{J}=\CC^+ \cup \CC^-$.
The optimal classifier is determined by the parameters $w$ and $b$ through solving the convex optimization problem %\cite{cortes1995support}:
\begin{subequations}\label{softmarginSVM}
\begin{align}  
              \min &\ \ {\frac{1}{2}\ \| w \|^2+C \sum_{i=1}^{l}\xi_i} \\
               \text{s.t.}&\ \               y_i(w^T\phi(x_i)+b)\geq1-\xi_i &  \;i = 1, \ldots, l\\
            & \ \  \xi_i\geq0  &     \;i = 1, \ldots, l
\end{align}
\end{subequations}
\noindent where $\phi$ maps training instances $x_i$ into a higher dimensional space, $\phi: \mathbb{R}^{n} \to \mathbb{R}^{m}$  ($m\geq n$). The term slack variables $\xi_i$ ($i \in \{1, \ldots, l\}$) in the objective function is used to penalize misclassified points. This approach is also known as {\em soft margin} SVM. The magnitude of penalization is controlled by the parameter $C$. Many existing algorithms (such as SMO, and its implementation in \cite{chang2011libsvm} that we use) solve the Lagrangian dual problem instead of the primal formulation, which is a popular strategy due to its faster and more reliable convergence.

% The Lagrangian dual of \ref{softmarginSVM} is formulated as:
%\begin{subequations}
% \begin{align}  
%              \max&\  \sum_{i=1}^{l}\alpha_i-\frac{1}{2}\sum_{i=1}^{l}\sum_{j=1}^{l}\alpha_i\alpha_j y_i y_jK(x_i,x_j) \\
%        \text{s.t.}&\ \               y_i(w^T\phi(x_i)+b)\geq1-\xi_i &        \;i = 1, \ldots, l\\
%            & \ \  \xi_i\geq0  &     \;i = 1, \ldots, l
%\end{align}
%\end{subequations}

The WSVM \cite{veropoulos1999controlling} (an extension of the SVM for imbalanced classes) assigns different weights to each data sample based on its importance, namely, %This provides a more flexible scheme compared to the regular SVM where the same penalization cost ($C$) is applied to all data points. 
%The formulation of WSVM for imbalanced binary classification is proposed by \cite{veropoulos1999controlling} using different costs associated with the positive ($C^+$) and negative ($C^-$) class: 
\begin{subequations}\label{eq:wsvm}
\begin{align}  
              \min &\ \ {\frac{1}{2}\ \| w \|^2+C^+ \sum_{\{i|y_i=+1\}}^{n_+}\xi_i+C^-\sum_{\{j|y_j=-1\}}^{n_-}\xi_j}\\
              \textrm{s.t.}&\ \  y_i(w^T\phi(x_i)+b)\geq 1-\xi_i \hspace{20pt}  i = 1, \ldots, l\\
&\ \ \xi_i\geq0 \hspace{100pt}  i = 1, \ldots, l
\end{align}
\end{subequations}
where the importance factors $C^+$, and $C^-$ are associated with the positive, and negative classes, and $|\CC^+|$ and $|\CC^-|$ are the sizes of majority and minority classes, respectively.
Problems \ref{softmarginSVM}, and \ref{eq:wsvm} can be transformed into the Lagrangian dual and solved using the Kuhn-Tucker conditions.

%\begin{subequations}\label{Lagrang}
%\begin{flalign}  
%              \max&\ \sum_{i=1}^{l}\alpha_i-\frac{1}{2}\sum_{i=1}^{l}\sum_{j=1}^{l}\alpha_i\alpha_j y_i y_jK(x_i,x_j) \\
%               \text{s.t.}& \ \        \sum_{j=1}^{l}\alpha_i y_i=0 &        \;i = 1, \ldots, l\\
%               \ \ 0\leq \alpha_i\leq C^+  &\  \textrm{if} \ \ y_i=+1  &  \textrm{and} \;i = 1, \ldots, l \\
%               \ \ 0\leq \alpha_i\leq C^-  & \   \textrm{if} \ \ y_i=-1  &\textrm{and}  \;i = 1, \ldots, l
%\end{flalign}
%\end{subequations}

The similarity between each pair of points $x_i$ and $x_j$ is measured by kernel function $k(\bx_i,\bx_j) = {\phi(x_i)}^T \phi(x_j)$. We use the Gaussian function (RBF) as the kernel function for the (W)SVM since Gaussian kernels typically achieve good performance under general smoothness assumptions \cite{tay2001application,xanthopoulos2014weighted} and are particularly well-suited if there is lack of additional knowledge of the data \cite{scholkopf2001learning}. 
Additional experiments with the polynomial kernel demonstrated much longer computational time while the achieved quality was very similar.
The RBF kernel is given by,
\begin{equation}
       k(\bx_i,\bx_j)=\exp(-\gamma \|\bx_i-\bx_j\|^2),  \gamma \geq 0.
 \end{equation} 
To achieve an acceptable quality of the classifier, many difficult data sets require reinforcement of (W)SVM with some parameter tuning methods. This tuning is called the model selection, and it is one of the most time-consuming components in (W)SVMs. In particular, tuning $C$, $C^+$, $C^-$, and kernel function parameters (e.g. bandwidth parameter $\gamma$ for RBF kernel function) can drastically change the quality of the classifier. 
In our solvers we employ an adapted nested uniform design model selection algorithm \cite{huang2007model}. It has been shown that the uniform design (UD) methodology for supervised learning model selection is more efficient and robust than other common methods, such as the grid search \cite{mangasarian2007privacy}. This method determines the close-to-optimal parameter set in an iterative nested manner. Since the test dataset might be imbalanced we select the optimal parameter set with respect to the maximum {\em G-mean} value.  

\section{Multilevel Support Vector Machines}
\begin{figure}
\includegraphics[width=0.47\textwidth]{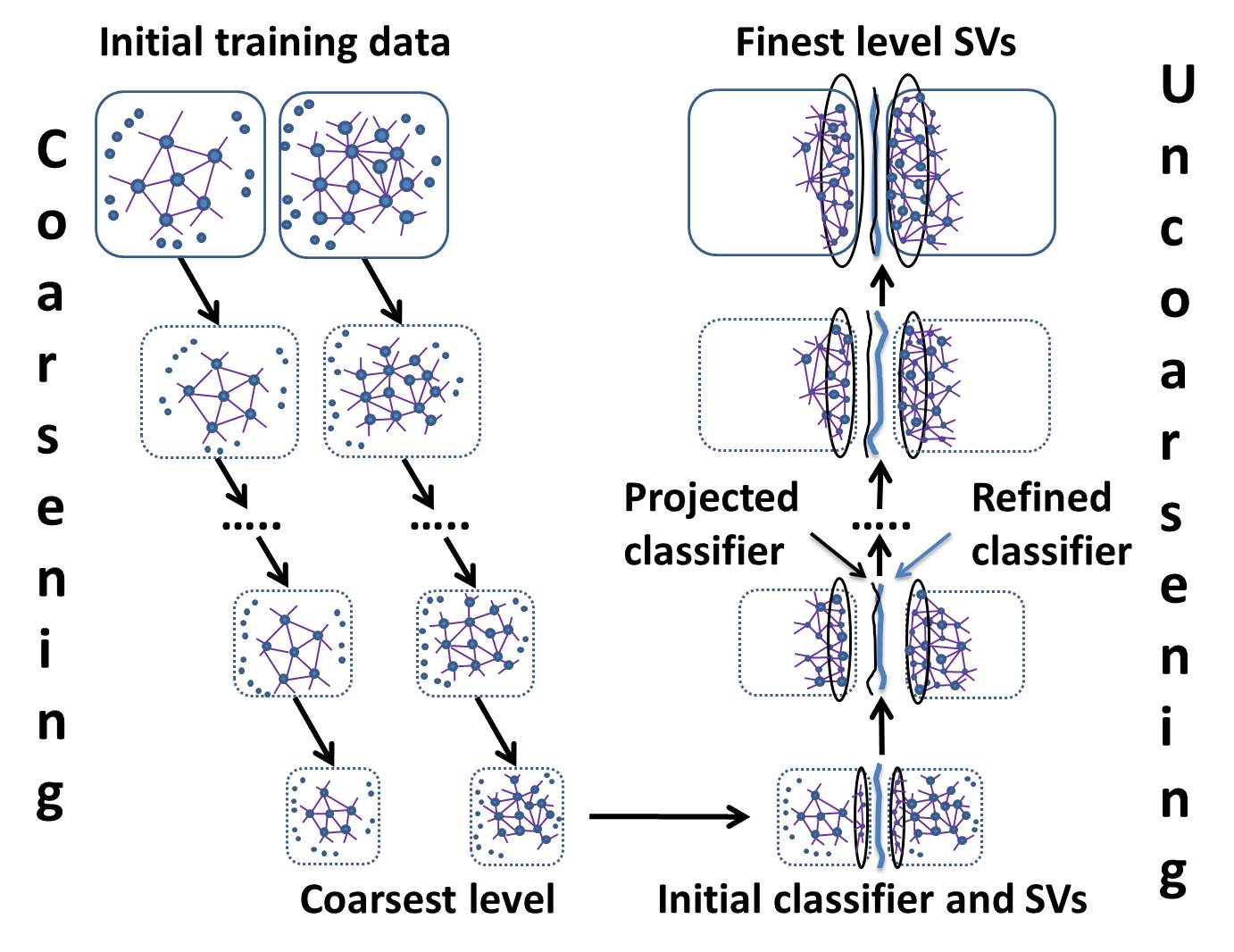}
\caption{The multilevel SVM framework.}\label{fig:ml}
\end{figure}
The multilevel framework (See Figure \ref{fig:ml}) proposed in this paper includes three phases: gradual training set  coarsening, coarsest support vectors' learning, and gradual support vectors' refinement (uncoarsening). Separate coarsening hierarchies are created for both classes $\CC^+$, and $\CC^-$, independently. Each next-coarser level contains a subset of points of the corresponding fine level. These subsets are selected using the approximated $k$-nearest neighbor graphs. In contrast to the coarsening used in multilevel dimensionality reduction method \cite{saaddim}, we found that selecting an \emph{independent} set only (including possible maximization of it) does not lead to the best computational results. Instead, making the coarsening less aggressive makes the framework much more robust to the changes in the parameters.
After the coarsest level is solved exactly, we gradually refine the support vectors and the corresponding classifiers level by level.
\begin{algorithm}
\caption{The Coarsening}
\label{alg1}
\begin{algorithmic}[1]
\State {\bf Input:} $G=(V,E)$ for class $\CC$
 \State  $\hat{V} \gets $ select maximal independent set in $G$
 \State   $\hat{U} \gets V \setminus \hat{V}$
\While{$\vert \hat{V}\vert<Q\cdot |V|$}

\While{$\hat{U} \neq \emptyset$}

     \State  randomly pick $i \in \hat{U}$
     \State $\hat{U} \gets \hat{U} \setminus \{i\}$ 
     \State $\hat{U} \gets \hat{U} \setminus N(i,\hat{U})$
     \State $\hat{V} \gets \hat{V} \cup \{i\}$ \Comment add point $i$ to coarse set
     
\EndWhile

  \State   $\hat{U} \gets V \setminus \hat{V}$  \Comment Reset the set of available points

\EndWhile\\
\Return $\hat{V}$
\end{algorithmic}
\end{algorithm}

%\begin{algorithm}
%\caption{The Coarsening}
%\label{alg1}
%\begin{algorithmic}[1]
% \State  $\hat{V} \gets \emptyset$
% \State   $\hat{U} \gets \emptyset$
% \State  Randomly Pick $t_0 \in V$; $T \gets T\cup \{t_0\}$.
%\While{$\vert \hat{V}\vert<Q\cdot n$}
%     \State  Randomly pick $i \in T$; $T \gets T\cup \{i\}$
%         \If {$i\notin \hat{U}\cup \hat{V}$}
%            \State $\hat{V}\gets \hat{V} \cup \{i\}$
%             \For{ $\forall (i,j) \in E,  j \notin \hat{U}$} 
%                 \State   $\hat{U} \gets \hat{U} \cup \{j\}$
%                    \For{$\forall (j,k) \in E$}
%                        \If {$k \notin \hat{U}\cup \hat{V}$}
%                           \State $T \gets  T \cup \{k\}$
%                       \EndIf
%                     \EndFor
%             \EndFor
%         \EndIf
%\EndWhile
%
%\end{algorithmic}
%\end{algorithm}

\subsection{The Coarsening Phase}
The coarsening algorithms are the same for both $\CC^+$, and $\CC^-$, so we provide only one of them. Given a class of data points $\CC$, the coarsening begins with a construction of an approximated $k$-nearest neighbors (A$k$NN) graph $G=(V,E)$, where $V=\CC$, and $E$ are the edges of A$k$NN. The goal is to select a set of points $\hat{V}$ for the next-coarser problem, where $\vert\hat{V}\vert \geq Q \vert V\vert$, and $Q$ is the parameter for the size of the coarse level graph (see Section \ref{sec:disc}). The second requirement for $\hat{V}$ is that it has to be a dominating set of $V$. 

The coarsening for class $\CC$ is presented in Algorithm \ref{alg1}. The algorithm consists of several iterations of independent set of $V$ selections that are complementary to already chosen sets. We begin with choosing a random independent set (line 2) using greedy algorithm. It is eliminated from the graph, and the next independent set is chosen and added to $\hat{V}$ (lines 5-10).
For imbalanced cases, when WSVM is used, we avoid of creating very small coarse problems for $\CC^-$. Instead, already very small class is continuously replicated across the rest of the hierarchy if $\CC^+$ still requires coarsening. We note that this method of coarsening will reduce the degree of skewness in the data and make the data approximately balanced at the coarsest level.

%\begin{algorithm}[H]
%
% \eIf {$\|n^+\| < thresh}$}{
%       Avoid  coarsening the minority class \;
%}{
%  $\hat{V} \longleftarrow \emptyset$  \;
%  $\hat{U} \longleftarrow \emptyset$  \;
%  Randomly Pick $t_0 \in V$; $T \gets T\cup \{t_0\}$.   \;
%\While{$\vert \hat{V}\vert<0.6*n$}{
%     Randomly pick $i \in T$; $T \gets T\cup \{i\}$   \;
%         \eIf {$i\notin \hat{U}\cup \hat{V}$}{
%             $\hat{V}\gets \hat{V} \cup \{i\}$  \;
%             \For{ $\forall (i,j) \in E,  j \notin \hat{U}$} {
%                   $\hat{U} \gets \hat{U} \cup \{j\}$  \;
%                    \For{$\forall (j,k) \in E$}{
%                        \eIf {$k \notin \hat{U}\cup \hat{V}$}{
%                            $T \longleftarrow T \cup \{k\}$} \;
%                          }
%                     }
%            }
%         }
%}
%}
%\caption{Coarsening by random independent set}
%\label{alg1}
%\end{algorithm}

%
%\begin{algorithm}
%\caption{Coarsening(Input: set $\JJ$)}
%\label{alg2}
%\begin{algorithmic}
%\If {$|\JJ| <$ $Ulim$}
%       \State Stop coarsening
%       \State Solve the coarsest (W)SVM exactly
%\Else
%     \If {$|\CC^+|>$ $Mlim$}
%       \State $\hat{V}_+ \leftarrow $ result of Algorithm \ref{alg1} on $\CC^+$
%    \Else
%    	   \State $\hat{V}_+\leftarrow V$
%    \EndIf
%     \If {$|\CC^-|>$ $Mlim$}
%       \State $\hat{V}_- \leftarrow $ result of Algorithm \ref{alg1} on $\CC^-$
%    \Else
%    	   \State $\hat{V}_-\leftarrow V$
%    \EndIf
%    
%
%\EndIf
%
%\end{algorithmic}
%\end{algorithm}

The multilevel framework recursively calls the coarsening process until it creates a hierarchy of $r$ coarse representations $\{\JJ_i\}_{i=0}^r$ of $\JJ$. At each level of this hierarchy, the corresponding A$k$NNs' $\{G_i=(V_i,E_i)\}_{i=0}^r$ are saved for future use at the uncoarsening phase. The corresponding data and labels at level $i$ is denoted by $(X_i,Y_i) \in \mathbb{R}^{k\times{(n+1)}}$, where $|X_i|=k$.

\subsection{The Coarsest Problem}

At the coarsest level $r$, when $|\JJ_r| << \JJ$, we can apply an exact algorithm for training the coarsest classifier. Typically, the size of the coarsest level depends on the computational resources. However, for the (W)SVM problems, one can also consider some criteria of the separability between $\CC_r^+$, and $\CC_r^-$ \cite{wang2008feature}, i.e., if a fast test exists or some nice data properties are available. We used the simplest criterion bounding $\JJ_r$ to 500. Processing the coarsest level includes an application of the UD \cite{huang2007model} model selection to get high-quality classifiers on the difficult data sets.

%A sequence of graphs $G_1,G_2,\ldots,G_r$  can be generated by recursively coarsening the initial graph $G_0=(V_0,E_0)$, where $G_i=(V_i,E_i)$ is the coarse graph of level $i$, $i=1,\ldots,r$, where $V_i \subset V_0$ is a set of indices of the given data $x_1,x_2,\ldots,x_k$, $k < l$.  The corresponding data and labels at level $i$, $i=1,\ldots,r$, is denoted by $(X_i,Y_i) \in \mathbb{R}^{k\times{(n+1)}}$ and the coarsest graph is represented by $G_r$.

\subsection{The Refinement Phase}
\label{refine}
Given the solution of coarse level $i+1$ (the set of support vectors $S_{i+1}$, and parameters $C_{i+1}$, and $\gamma_{i+1}$), the primary goal of the refinement is to update and optimize this solution for the current fine level $i$. Unlike many other multilevel algorithms, in which the inherited coarse solution contains projected variables only, in our case, we initially inherit not only the coarse support vectors (the solution that can represent the whole training set \cite{syed1999incremental,fung2002incremental}) but also parameters for model selection. This is because the model selection is an extremely time-consuming component of (W)SVM, and can be prohibitive at fine levels. However, at the coarse levels, when the problem is much smaller than the original, we can apply much heavier methods for model selection without any loss in the total complexity of the framework.

\begin{figure}
\centering
\includegraphics[width=0.3\textwidth]{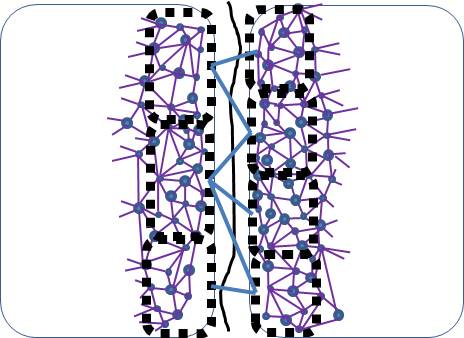}
\caption{Refinement of clusters of SVs and their neighbors. Dotted lines correspond to clusters. Pairs of connected clusters, whose centroids are neighbors, are re-trained, and their support vectors are updated. }\label{fig:clust}
\end{figure}

%For this phase, we first classify the data points at the level $r$th (coarsest) using the SVM and WSVM algorithms and perform the UD model selection on the training data. For level $i$th, $i=r-1,r-2,\ldots,0$, we implement the SVM and WSVM algorithms, initialized by the support vectors at $(i+1)$st level. Previous studies have shown that the classification characteristics of the whole training set can be represented by support vectors \cite{syed1999incremental,fung2002incremental}. 
%The state-of-art studies have developed several incremental learning methods to update support vectors each time new data are added to the available dataset \cite{cauwenberghs2001incremental,syed1999incremental,fung2002incremental}.
\small
\begin{algorithm}
\caption{The Refinement at level $i$}
\label{alg3}
\begin{algorithmic}[1]
\State {\bf Input:} $\JJ_i, S_{i+1}, C_{i+1}, \gamma_{i+1}$
% \State  Given   $\mathcal{J}=\{(x_i, y_i)\}_{i=1}^l$  classify it into two classes.
%Given $X=[x_1,x_2,\ldots,x_l] \in \mathbb{R}^{l\times{n}}$ and $Y=[y_1,y_2,\ldots,y_l] \in \mathbb{R}^{l\times{1}}$, classify it into two classes.
\If {$i$ is the coarsest level}
 \State Calculate the best ($C_{i}$, $ \gamma_{i}$) using UD
  \State $S_i \leftarrow$ Apply SVM on $X_i$
%  \State {\bf Return} $S_i$, $C_i$, $\gamma_i$
  \EndIf
%  \State $C_r \gets C_{optimal}$
%  \State $\gamma_r \gets \gamma_{optimal}$.
   
        %            \For{$i = r-1,\ldots,0$}
                          \State Calculate nearest neighbors $N_i$ for support vectors $S_{i+1}$ from the existing A$k$NN $G_i$ 
                           \State  ${data}^{(i)}_{train} \gets S_{(i+1)} \cup N_i$
                                 \If {$|{data}^{(i)}_{train}| < Q_{dt}$ (see Section \ref{sec:disc})}
                                       \State $C^O  \gets C_{i+1}$; $\gamma^{O} \gets \gamma_{i+1}$
                                        \State Run UD using the initial center $(C^O, \gamma^{O})$
                                       
                                 \Else
                                       \State $C_i \leftarrow C_{i+1}$; $\gamma_i \leftarrow \gamma_{i+1}$%Inherit the parameter $C_{i+1}$ and $\gamma_{i+1}$ to level $i$th.
                                  \EndIf
                                  \If {$|{data}^{(i)}_{train}| \geq Q_{dt}$ }\par
                                 \State Cluster ${data}^{(i)}_{train}$ into $K$ clusters
                                 \State $\forall k\in K$ find $P$ nearest opposite-class clusters
                                % \State and  train each subset of two nearest neighbor 
                               %  \State class (minority  and majority)
%                                 \State Find the class label of each test instance
                             %    \State based on its closest sum distance to  the 
                               %   \State center of clusters (of minority and majority
                              %    \State  classes)
               \State $S_i \leftarrow$ Apply SVM on pairs of nearest clusters only

                                  \Else
                                       \State $S_i \leftarrow$ Apply SVM directly on ${data}^{(i)}_{train}$
                                   %     \State  initialized by the support vectors at level $(i+1)$.
                                  \EndIf
                                  \State {\bf Return} $S_i$, $C_i$, $\gamma_i$
         %            \EndFor
\end{algorithmic}
\end{algorithm}
\normalsize
The refinement is presented in Algorithm \ref{alg3}. The coarsest level is solved exactly and reinforced by the model selection (lines 2-5). If $i$ is one of the intermediate levels, we build the set of training data $data^{(i)}_{train}$ by inheriting the coarse support vectors $S_{i+1}$ and adding to them some of their approximated nearest neighbors at level $i$ (lines 6-7) (in our experiments, usually not more than 5). If the size of $data^{(i)}_{train}$ is still small (relatively to the existing computational resources, and the initial size of the data) for applying model selection, and solving SVM on the whole $data^{(i)}_{train}$, then we use coarse parameters $C_{i+1}$, and $\gamma_{i+1}$ as initializers for the current level, and retrain (lines 9-10,19). Otherwise, the coarse $C_{i+1}$, and $\gamma_{i+1}$ are inherited in $C_{i}$, and $\gamma_{i}$ (line 12). Then, being large for direct application of the SVM, $data^{(i)}_{train}$ is clustered into several clusters, and pairs of nearest opposite clusters are retrained, and contribute their solutions to $S_i$ (lines 15-17, see Figure \ref{fig:clust}). \emph{We note that cluster-based retraining can be done in parallel, as different pairs of clusters are independent. Moreover, the total complexity of the algorithm does not suffer from reinforcing the cluster-based retraining with model selection.}

\subsection{Performance Measures}

Evaluation of the proposed algorithm is done using the classification confusion matrix. For binary classification, the values 

%\footnotesize
%\begin{wraptable}{r}{3cm}
%\vspace{-15pt}
%\caption*{\emph{Confusion matrix.}}\label{wrap-tab:1}
%\vspace{-10pt}
%\begin{tabular}{|c|c|c|}
%\hline & $\CC^+$ & $\CC^-$\\
%\hline $\CC^+$ & TP & FP \\
%\hline $\CC^-$ & FN & TN \\\hline 
%\end{tabular}
%\vspace{-10pt}
%\end{wraptable} 
%\normalsize

%\footnotesize
\begin{table}
\caption{\emph{Confusion matrix.}}\label{wrap-tab:1}
\centering
\begin{tabular}{|c|c|c|}
\hline & $\CC^+$ & $\CC^-$\\
\hline $\CC^+$ & $TP$ & $FP$ \\
\hline $\CC^-$ & $FN$ & $TN$ \\\hline 
\end{tabular}
\end{table} 
\normalsize

\noindent $TP$, $FP$, $FN$, $TN$ denote true positives, false positives, false negatives and true negatives, respectively. The higher $TP$ and $TN$ indicate a better classifier.
Accuracy (ACC) is one of most common performance measure for classification, which is defined as the correctly classified test data over the whole training points. However, the accuracy measure might not be a proper performance measure for imbalanced problems, since the performance of this metric is mostly dominated by the majority class. Therefore, we use sensitivity (SN) and specificity (SP) defined as
\begin{equation}
\textrm{SN}=\frac{TP}{TP+FN}, \ \ \textrm{SP}=\frac{TN}{TN+FP}.
\end{equation}
Another well-known metric is the geometric mean of sensitivity and specificity (G-mean) which is defined as:
\begin{equation}
\textrm{G-mean}=\sqrt{\textrm{Specificity}*\textrm{Sensitivity}}
\end{equation}

%We report sensitivity, specificity, {\em G-mean}, and accuracy in this paper. 

\section{Computational Results}

Our multilevel SVM and WSVM are evaluated on binary classification benchmarks at UCI \cite{UCI}, and the cod-rna dataset \cite{alon1999broad}.
Single SVM and WSVM models are solved using LIBSVM-3.18 \cite{chang2011libsvm}, and the A$k$NN graphs are costructed using FLANN library \cite{muja_flann_2009}. Multilevel frameworks are implemented in MATLAB 2012a, and evaluated on Linux. The implementation is non-parallel and has not been optimized. The results for multilevel (W)SVM (objectives and running times) should only be considered qualitatively and can certainly be further improved by a more advanced  implementation. The implementation is available at \url{http://www.cs.clemson.edu/~isafro/software.html}.

We normalize all data prior to classification in order to get zero mean and unitary standard deviation. We perform a 9- and 5-point run design for the first and second stages of the nested UD, which has been demonstrated to be appropriate for the UCI data \cite{huang2007model}. The performance measures, including SN, SP, G-mean, and accuracy, are reported using the test data.

\begin{table}
\footnotesize
\centering
\caption{Benchmark data sets.}
\label{table1} 
    \begin{tabular}{cccccccc}         %|c|c|c|c|c|c|c|c|
    
        \hline
Dataset	&	$r_{imb}$	&	$n_f$ &	$|\JJ|$	&	$|\CC^-|$	&	$|\CC^+|$	\\ \hline
Letter26	&	0.96	&	16	&	20000	&	734	&	19266	\\
Ringnorm	&	0.50	&	20	&	7400	&	3664	&	3736	\\
Buzz &	0.80	&	77	&	140707	&	27775	&	112932	\\
Clean (Musk)	&	0.85	&	166	&	6598	&	1017	&	5581	\\
Advertisement	&	0.86	&	1558	&	3279	&	459	&	2820	\\
ISOLET	&	0.96	&	617	&	6238	&	240	&	5998	\\
cod-rna	&	0.67	&	8	&	59535	&	19845	&	39690	\\
Twonorm	&	0.50	&	20	&	7400	&	3703	&	3697	\\
Nursery	&	0.67	&	8	&	12960	&	4320	&	8640	\\
EEG Eye State	&	0.55	&	14	&	14980	&	6723	&	8257	\\
Hypothyroid	&	0.94	&	21	&	3919	&	240	&	3679	\\
\hline
\end{tabular}
\end{table}
\normalsize 

We assign the weights of each class proportional to the inverse of the class size, $\frac{C}{2|\CC^+|}$ and $\frac{C}{2|\CC^-|}$. This weighting method has been applied in a number of studies \cite{liu2005new,du2005weighted,huang2005weighted}. Clearly, the weights of two classes approximately equal when the data is balanced.

We evaluate the proposed method for both balanced and imbalanced problems. The experimental data have different imbalance ratios $r_{imb} = \frac{|\CC^+|}{|\JJ|}$. The details of the datasets are described in Table \ref{table1}.

The performance measures of the multilevel SVM (Table \ref{table2}) and multilevel WSVM (Table \ref{table3}) are compared with the regular (one-level) SVM (Table \ref{table4}) and WSVM (Table \ref{table5}). Since several components in the coarsening, and uncoarsening schemes are randomized algorithms (choosing coarse variables, A$k$NN, and clustering), the average numbers over  100 random runs are reported for each data set. We do not report the standard deviations because in all experiments they are negligibly small. Bold fonts emphasize the best results for the particular data sets.

\begin{table}[h]
\footnotesize
\centering
\caption{Performance measures for Multilevel SVM. Each cell contains an average over 100 executions. Column 'Depth' shows the number of levels in the multilevel hierarchy.}
\label{table2} 
    \begin{tabular}{cccccc}                  %|c|c|c|c|c|c|
        \hline
% \multirow{2}[0]{*}{Dataset}                         & \multicolumn{4}{c}{SVM}  &        \\\cline{2-6}
Dataset	&	ACC	&	SN	&	SP	&	G-mean	&	Depth	\\\hline
Letter$26$	&	$0.98   $	&	$0.99   $	&	$0.95   $	&	$0.97$	&	8	\\
Ringnorm	&	$0.98   $	&	$0.98  $	&	$0.99  $	&	\textbf{0.98}  	&	6	\\
Buzz 	&	$0.94   $	&	$0.96  $	&	$0.85  $	&	0.90 $ $	& 14		\\
Clean (Musk)	&	$1.00   $	&	$1.00  $	&	$0.99   $	&	\textbf{0.99} 	&	5	\\
Advertisement	&	$0.94$	&	$0.97$	&	$0.79  $	&	$0.87$   &	7	\\
ISOLET	&	$0.99   $	&	$1.00   $	&	$0.83  $	&	\textbf{0.92} 	&	11	\\
cod-rna	&	$0.95  $	&	$0.93   $	&	$0.97  $	&	$0.95  $	 &	9	\\
Twonorm	&	$0.97   $	&	$0.98   $	&	0.97 	&	\textbf{0.97}  		&	6	\\
Nursery	&	$1.00  $	&	$0.99   $	&	$0.98  $	&	\textbf{0.99} 	&	10	\\
EEG Eye State	&	$0.83  $	&	$0.82  $	&	$0.88   $	&	$0.85$	&	6	\\
Hypothyroid	&	$0.98   $	&	$0.98   $	&	$0.74  $	&	$0.85$	&	4	\\
\hline
\end{tabular}
\end{table}
\normalsize

\begin{table}[h]
\footnotesize
\centering
\caption{Performance measures for regular SVM.}
\label{table4} 
    \begin{tabular}{ccccccccc}                %|c|c|c|c|c|c|c|c|c|
        \hline
%\multirow{2}[0]{*}{Dataset}                            & \multicolumn{4}{|c|}{SVM}  \\\cline{2-5}
Dataset	&	ACC	&	SN	&	SP	&	G-mean				\\\hline
Letter$26$	&	1.00   	&	1.00   	&	0.97   	&	0.98  	\\
Ringnorm	&	0.98   	&	0.99  	&	0.98   	&	0.98   	\\
Buzz &	0.97  	&	0.99	  &	0.81  	&	0.89  	\\
Clean (Musk)	&	1.00    	&	1.00  	&	0.98  	&	0.99   		\\
Advertisement	&	0.92  	&	0.99   	&	0.45  	&	0.67 		\\
ISOLET	&	0.99   	&	1.00   	&	0.85   	&	0.92  		\\
cod-rna	&	0.96   	&	0.96  	&	0.95 	&	0.96   	\\
Twonorm	&	0.98  	&	0.98   	&	0.99   	&	0.98   	\\
Nursery	&	1.00   	&	1.00   	&	1.00   	&	1.00  		\\
EEG Eye State	&	0.88   	&	0.90  	&	0.86   	&	0.88   		\\
Hypothyroid	&	0.99  	&	1.00   	&	0.71  	&	0.83  		\\
\hline
\end{tabular}
\normalsize
\end{table}

Tables \ref{table2}, and \ref{table4} demonstrate that the quality of the proposed multilevel SVM algorithms is very similar to the quality of the single-level SVM algorithm. Similar conclusion (Tables \ref{table3}, and \ref{table5}) can be stated about the comparison of multilevel WSVM, and the regular single-level WSVM, except the experiments with the Advertisement dataset for which the multilevel WSVM is significantly better than the single-level SVM.
Multilevel WSVM usually performs better than Multileve SVM in cases when the data is imbalanced. Both multilevel frameworks will decrease the imbalance ratio in the coarsest level and can make the training data approximately balanced.

%In Table \ref{table4} the computational time is presented. It is clear from the boldface results under the multilevel SVM-based algorithms, that the multilevel SVM-based algorithms perform  much faster than the single-level methods  for all datasets in general. 
\emph{The main achievement of the proposed multilevel scheme is its computational time (see Table \ref{table6}) that substantially improves that of the single-level (W)SVM when the model selection techniques must be applied on difficult data sets. We note that for most of the datasets in the benchmark, using model selection was extremely important for obtaining high-quality results. Moreover, the observed improvement is not complete, because (similar to many multilevel and multigrid algorithms) the refinement phase can be easily parallelized at levels where the training by clusters is employed.}

\begin{table}
\footnotesize
\centering
\caption{Performance measures for Multilevel WSVM. Each cell contains an average over 100 executions. Column 'Depth' shows the number of levels in the multilevel hierarchy.}
\label{table3} 
    \begin{tabular}{cccccccccc}               %|c|c|c|c|c|c|c|c|c|c|
        \hline
%\multirow{2}[0]{*}{Dataset}                          &  \multicolumn{4}{c}{WSVM} &\\\cline{2-6}
Dataset	&	ACC	&	SN	&	SP	&	G-mean	&	Depth	\\\hline
Letter$26$	&	$0.99   $ 	&	$0.99   $ 	&	$0.96  $	&	\textbf{0.99}  $ $	&	8	\\
Ringnorm	&	$0.98   	$ &	$0.97   $ 	&	$0.99   $	&	\textbf{0.98}  $ $	&	6	\\
Buzz	&	$0.94   	$ &	$0.96   $ 	&	$0.87   $	&	\textbf{0.91}  $ $	& 14		\\
Clean (Musk)	&	$1.00   $	&	$1.00   $	&	$0.99  $	&	\textbf{0.99}  $ $	&	5	\\
Advertisement	&	$0.94	   $&	$0.96	 8 $ &	$0.80	   $ &	\textbf{0.88} $ $	&	7	\\
ISOLET	&	$0.99  $	&	$1.00  $	&	$0.85   $	&	\textbf{0.92} $ $	&	11	\\
cod-rna		&	$0.94  $	&	$0.97  $	&	$0.95  $	&	\textbf{0.96}  $ $	&	9	\\
Twonorm	&	$0.97   $	&	$0.98   $	&	$0.97   $	&	\textbf{0.97}  $ $	&	6	\\
Nursery		&	$1.00  $	&	$0.99  $	&	$0.98   $	&	\textbf{0.99} $ $	&	10	\\
EEG Eye State	&	$0.87  $	&	$0.89  $	&	$0.86  $	&	\textbf{0.88} $ $	&	6	\\
Hypothyroid		&	$0.98  $	&	$0.98  $	&	$0.75  $	&	\textbf{0.86} $ $	&	4	\\
\hline
\end{tabular}
\normalsize
\end{table}

\begin{table}
\footnotesize
\centering
\caption{Performance measures for regular WSVM.}
\label{table5} 
    \begin{tabular}{ccccccccc}                %|c|c|c|c|c|c|c|c|c|
        \hline
%  \multirow{2}[0]{*}{Dataset}                       & \multicolumn{4}{|c|}{WSVM} \\\cline{2-5}	
Dataset	&	ACC	&	SN	&	SP	&	G-mean		\\\hline
Letter$26$	&	1.00   	&	1.00   	&	0.97   	&	0.99   	\\
Ringnorm		&	0.98  	&	0.99   	&	0.98  	&	0.98   	\\
Buzz 	&	0.96  	&	0.99 	&	0.81 	&	0.89  	\\
Clean (Musk)	&	1.00  	&	1.00   	&	0.98    	&	0.99   	\\
Advertisement		&	0.92  	&	0.99  	&	0.45   	&	0.67   	\\
ISOLET	&	0.99   	&	1.00   	&	0.85   	&	0.92   	\\
cod-rna		&	0.96   	&	0.96   	&	0.96   	&	0.96   	\\
Twonorm	&	0.98   	&	0.98   	&	0.99   	&	0.98   	\\
Nursery	&	1.00   	&	1.00   	&	1.00  	&	1.00  	\\
EEG Eye State	&	0.88   	&	0.90   	&	0.86   	&	0.88   	\\
Hypothyroid		&	0.99  	&	1.00   	&	0.75   	&	0.86   	\\
\hline
\end{tabular}
\normalsize
\end{table}

\begin{table}
\footnotesize
\centering
\caption{Computational time (in sec.)}
\label{table6} 
    \begin{tabular}{|c|c|c|c|c|}
        \hline
        \multirow{4}[0]{*}{Dataset}                 & \multicolumn{4}{|c|}{Multilevel}\\\cline{2-5}
                      & \multicolumn{2}{|c|}{Yes}  & \multicolumn{2}{|c|}{No}   \\\cline{2-5}	
	&	 \multicolumn{2}{|c|}{ModelSelection} 	&	 \multicolumn{2}{|c|}{ModelSelection} 	\\\cline{2-5}				
        	&	Yes	&	No	&	Yes	&	No	\\\hline
Letter26	&	\bf{45}	&	112	&	333	&	27	\\
Ringnorm	&	\bf{4}	&	21	&	42	&	12	\\
Twonorm	&	\bf{4}	&	21	&	45	&	12	\\
Buzz &	\bf{2329}	&	\bf{2400}	&	70452	&	20386	\\
Clean (Musk)	&	\bf{30}	&	92	&	167	&	55	\\
Advertisement	&	\bf{196}	&	104	&	412	&	41	\\
ISOLET	&	\bf{69}	&	373	&	1367	&	297	\\
cod-rna	&	\bf{172}	&	293	&	1611	&	208	\\
Nursery	&	\bf{63}	&	37	&	519	&	42	\\
EEG Eye State	&	\bf{51}	&	32	&	447	&	33	\\
Hypothyroid	&	\bf{3}	&	3	&	5	&	1	\\
\hline
\end{tabular}
\normalsize
\end{table}

\section{Discussion}\label{sec:disc}
\paragraph{Complexity of the multilevel framework.} The complexity of the multilevel (W)SVM framework is determined by the complexities of coarsening, and uncoarsening. The complexity of coarsening is similar to that of the very many different multilevel algorithms such as \cite{RonSB11,saaddim,brandt:optstrat,abou2006multilevel}. The most time-consuming component of it is A$k$NN construction which is linear in $|\JJ|$. We experimented with exact algorithms for finding $k$NN graphs, and no significant change in the quality was observed while the running time was much bigger. The complexity of the uncoarsening is one of the most beneficial parts of the framework as it is much faster than the typical linear-time  uncoarsening of multilevel algorithms. Usually, in multilevel and multigrid algorithms, all variables are refined when level $i+1$ is uncoarsened to level $i$. In our framework, we do not need this as we work with the inherited set of the support vectors and their limited number of neighbors. In our experiments, we used 5 approximate nearest neighbors. Clustering of large sets of $|S_{i+1} \cup N_i|$ was done using fast k-means. We note, that it is not important to cluster them precisely up to a very small error, and the algorithm is not sensitive to this at all. 

\paragraph{How sensitive are the parameters?} Similar to the multigrid, the multilevel frameworks require several parameters. However, the performance of the proposed framework is not too sensitive to most of them. 

\begin{figure}
%\vspace{-0.75cm}
 \centering
  \begin{tabular}{cc}
          \includegraphics[scale=0.2]{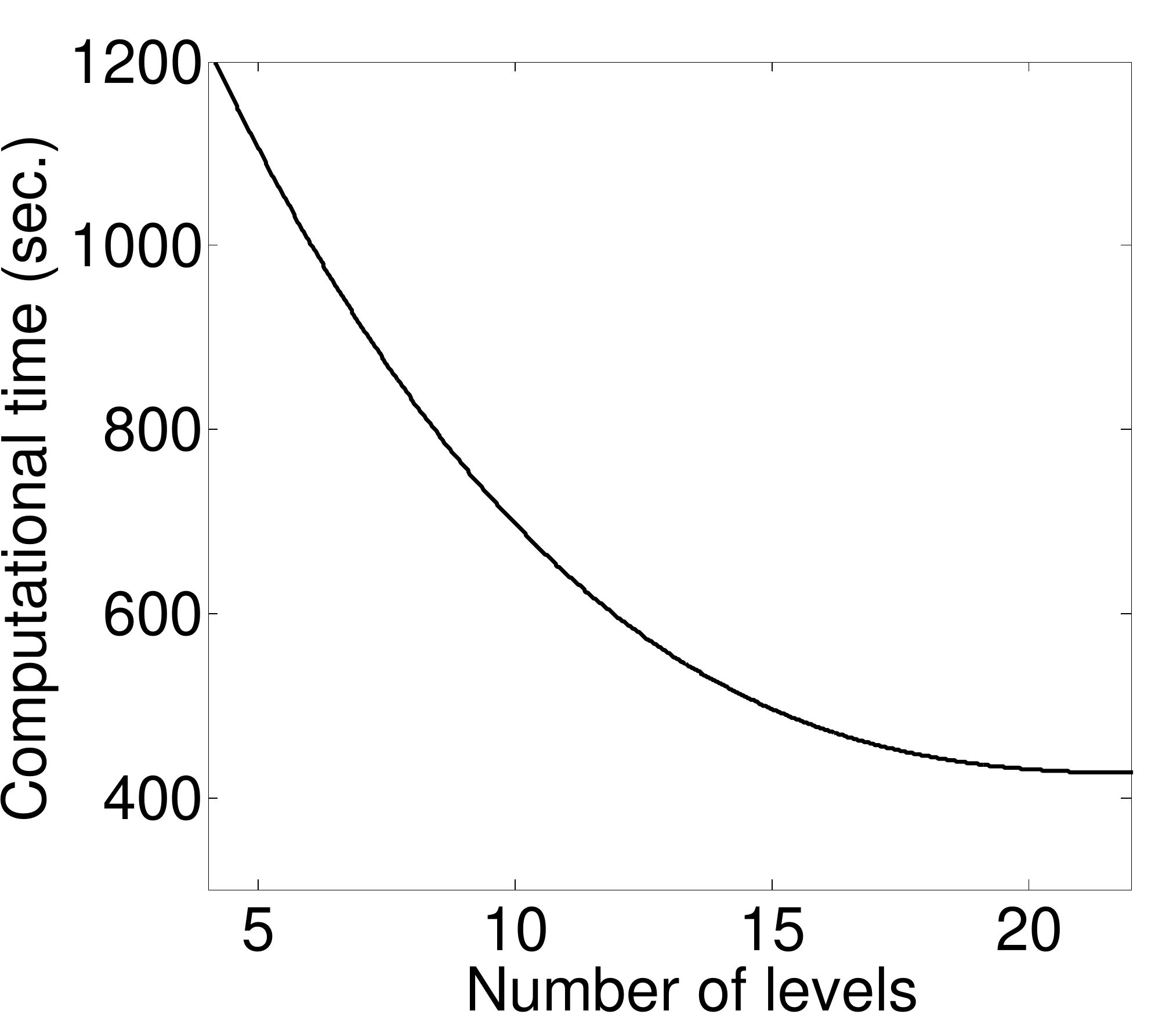} & \includegraphics[scale=0.2]{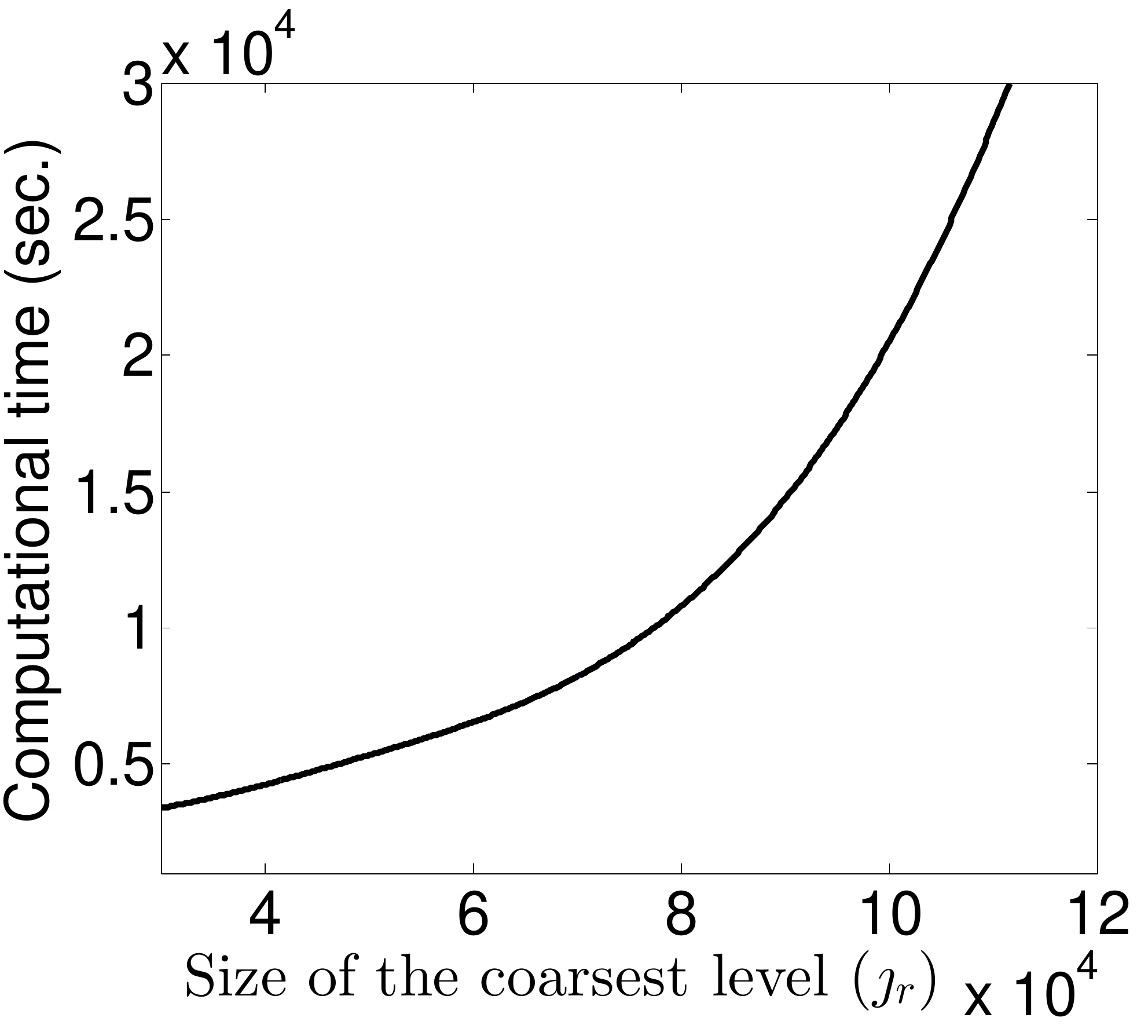}\\
                    (a) cod-rna                                           &           (b)  Buzz\\
%Computional time vs. the number of levels (Cod-rna dataset) &  Computional time vs. the size of the coarsest level ($\JJ$) (Buzz in social media dataset)
\end{tabular}
\caption{These two figures show computational time vs. the number of levels for cod-rna dataset and computional time vs. the size of the coarsest level ($|\JJ_r|$) for Buzz dataset, respectively}
\label{fig_CompLevel} 
\end{figure}

\noindent \emph{Size of the coarsest level, $|\JJ_r|$, or when to stop the coarsening?} If no additional knowledge about the data properties exist, the simplest way to choose this parameter is usually determined by the available computational resources. The greater $|\JJ_r|$, the better approximation will be prolongated throughout the uncoarsening. As this parameter decreases, the number of levels increases.  Figure \ref{fig_CompLevel}(a) shows that the algorithm performs three times faster for the cod-rna dataset, as the number of levels increases. This is similar to Buzz daset (see Figure \ref{fig_CompLevel}(b)). For large datasets, the algorithm performs slightly faster with more levels. 

\noindent \emph{Size of the next-coarser level ($Q$ in Algorithm \ref{alg1}).} Typically, too aggressive coarsening, i.e., very small $Q$, can lead to fast loss of information as the data is actually ``compressed'' by the coarsening. The framework, however, will run faster. In the presented results, we used $Q=0.6$. However, experiments with $Q=0.4, 0.5, 0.7$ do not significantly change the results. For larger $Q$, the coarsening becomes more gradual, and, thus, the running time of the framework is slowed down.

\noindent \emph{When to refine the entire set $S_{i+1} \cup N_i$, and when to cluster it, and refine by clusters ($Q_{dt}$ in Algorithm \ref{alg3})?} Similar to $|\JJ_r|$, it depends on the available computational resources, and on $|S_{i+1} \cup N_i|$. In our experiments we used $|S_{i+1} \cup N_i|=1000$ for larger data sets, and tried to decrease it to 500 for smaller.

\begin{figure}[h]
 \centering
  \begin{tabular}{cc}
          \includegraphics[scale=0.2]{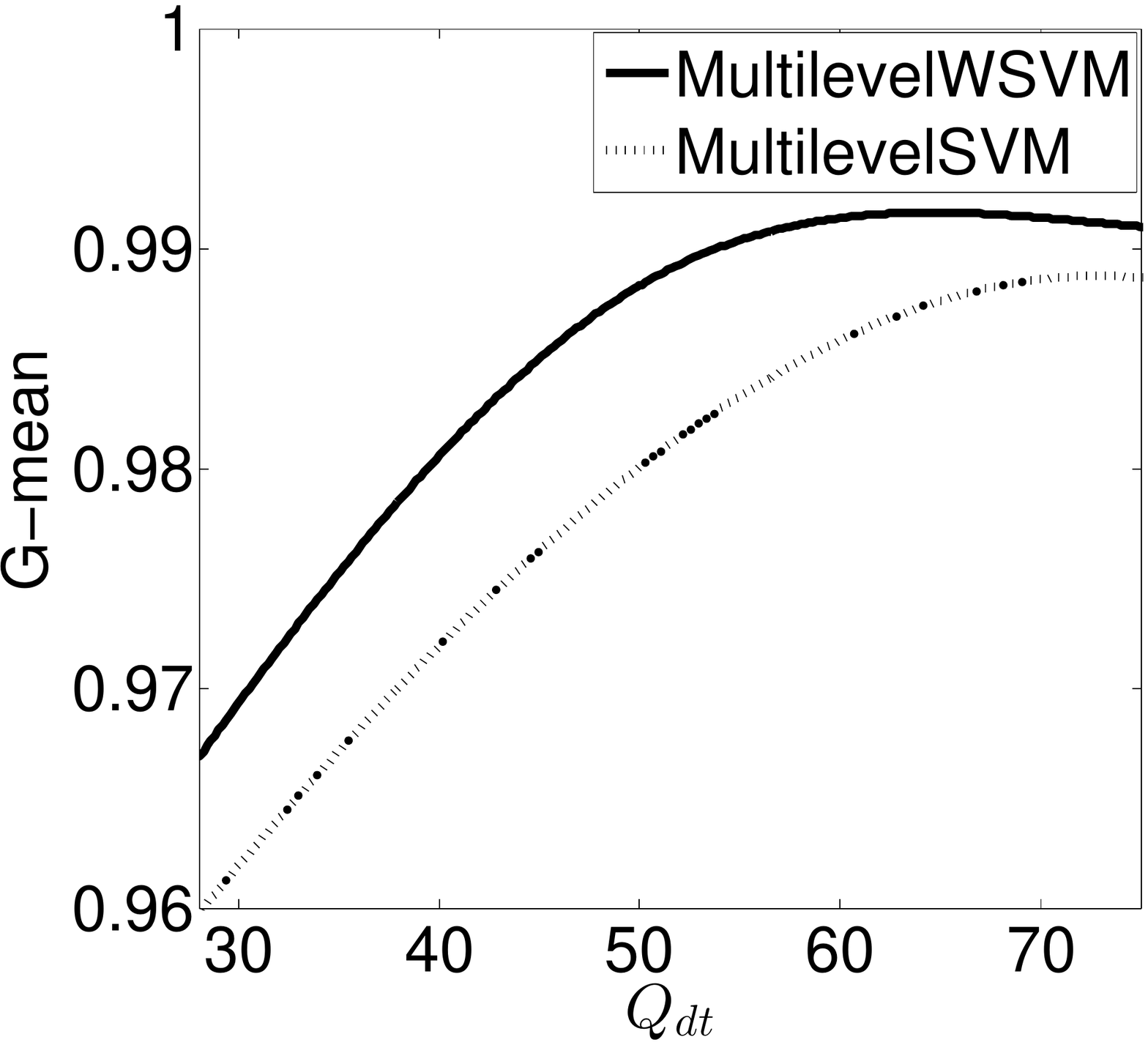} &\includegraphics[scale=0.2]{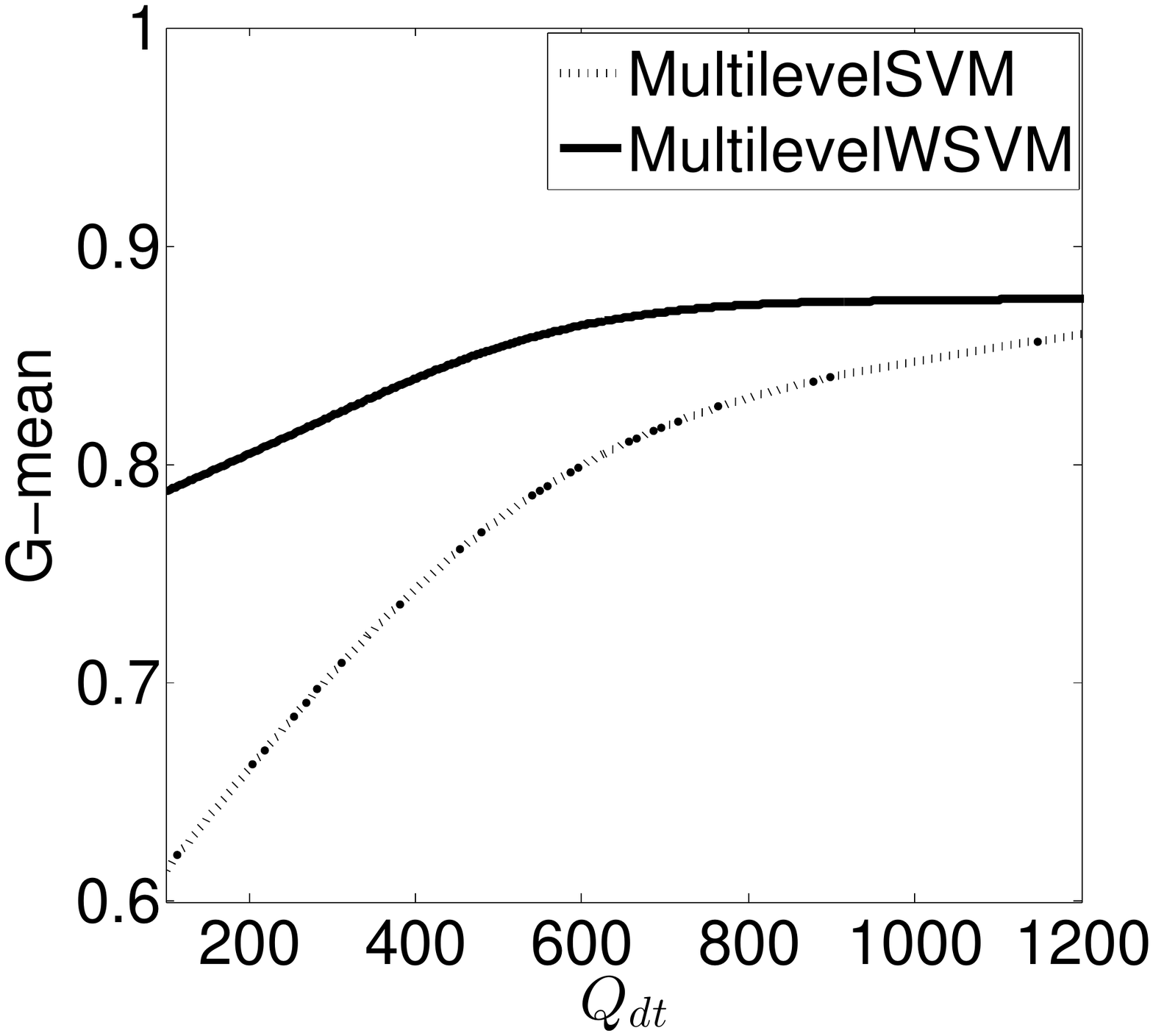} \\
            (a)  Letter26 &  (b) Hypothyroid 
\end{tabular}
\caption{Sensitivity of Multilevel SVM and Multilevel WSVM in terms of G-mean metric to $Q_{dt}$ for Letter26 data and Hypothyroid data}
\label{fig_parm_ModelS} 

\end{figure}

\noindent \emph{When to apply model selection?} The model selection component can be the most sensitive on the difficult data sets, on which achieving high-quality classifiers is nearly impossible with regular (W)SVM. One of the main advantages of the proposed multilevel framework is that it allows to apply the model selection at coarse levels only, when the size of the data is still small. Then, we stop to use model selection but the parameters $C$, and $\gamma$ are inherited from the coarse levels and serve as initializers at fine levels. The main drawback of the existing model selection methods is that searching for optimal parameters might become considerably computationally expensive for even not very large datasets, and existing methods lack efficient algorithmic ideas for improving it. On the other hand, running the model selection in a multilevel framework on pairs of very small clusters could yield sometimes poor results, because ``too optimized'' parameters for small clusters can be incorrect for a global solution. For simplicity we use the same $Q_{dt}$ to control this, while the sizes of clusters were 300. For all datasets, we also experimented  with different $Q_{dt}$. Clearly, increasing it improves the quality of the classifier. Figure \ref{fig_parm_ModelS}(a) shows that the performance measure decreases 5\% when this parameter decreases 20 times for the Letter26 data. The sensitivity of hypothyroid data to model selection parameter demonstrated in Figure \ref{fig_parm_ModelS}(b).

%The G-mean decreases between 12-30 \% for Hypothyroid data as this parameter desecrates from 3000 to 100. 
 
\paragraph{Future research and possible improvements.}

To the best of our knowledge the proposed method is the first multilevel algorithmic framework for (W)SVM. We identify several promising research direction that can potentially take such frameworks to the next level of quality. The most important of them are (a) better coarsening schemes, (b) prolongation of the parameters $C$, and $\gamma$ in the uncoarsening, and (c) merging support vectors in the cluster-based refinement.
\begin{itemize}
\item \emph{Coarsening.} In our framework, we coarsen two classes $\CC^+$, and $\CC^-$ independently of each other, except preserving the size of the coarse minority class, and making it balanced with the majority class at the very coarse levels. Separating two coarsenings can potentially lead to the problem of choosing the coarse variables that do not reflect well the boundaries between two classes, i.e., they can be selected far from the separating hyperplane, and, thus, more refinement is needed to train the best support vectors. One possible way to cope with such problem is to design a mutual coarsening for both classes. However, for such clustering, we need to define a new class for the ``boundary'' variables, i.e., for those that aggregate in themselves variables of both $\CC^+$, and $\CC^-$.
\item \emph{Prolongation of $C$, and $\gamma$.} Currently, when the refinement divides $data^i_{train}$ into clusters for re-training, we inherit both parameters from the coarse level. This helps the algorithm to initialize a better model. However, it is not straightforward to further update both $C$, and $\gamma$ in order to prolongate them to the next-finer level. In the current algorithm, we prolongate them directly without updating, if cluster-based refinement is applied.
\item \emph{Merging support vectors.} Although, it was not observed frequently in the experiments, a potential drawback of the cluster-based refinement is a large number of the support vectors, as some of them may play similar roles in the separation when chosen from different pairs of clusters. Thus, finding a fast strategy to merge them can potentially improve both quality and complexity of the multilevel framework. Another potential problem with the clustering-based refinement is too unbalanced clusters, when the clustering method assigns too few points in some clusters for re-training. This can create bias in the  classification.
\end{itemize}

%\begin{figure}[htbp]
%\centering
%          \includegraphics[scale=0.4]{Time_Uplim.pdf}
%\caption{Computional time vs. $Ulim$ (Buzz in social media dataset)}
%\label{fig_Uplim} 
%\end{figure*}

%\begin{figure}[htbp]
%\centering
%          \includegraphics[scale=0.4]{ParamModelSelechypo3.pdf}
%\caption{Sensitivity of Multilevel SVM and Multilevel WSVM in terms of G-mean metric to $P_{(MlS)}$ for Hypothyroid data}
%\label{fig_parm_ModelS_Hypo} 
%\end{figure}

%Parallel implementations on the bagging approach can even make the analysis problem much faster.

\section{Conclusion}

In this paper, we propose a novel fast multilevel (W)SVM-based algorithm for classification of large datasets. Experimental results on benchmark datasets demonstrate very promising results with no loss of quality in the performance measures. The reduced computational time provided by the multilevel framework is significant for large-scale data sets. In addition, the proposed methodology is very successful for large imbalanced classification problems since it can reduce the degree of skewness in the data and make the data approximately balanced at the coarse levels. 

%This study can be extended in several directions. We will improve the coarsening method other than randomly chosen independent set. The k-means clustering technique should be improved to avoid extreme imbalanced cases. To make the code faster, learning on pairs of clusters can be parallelized due to their independance. Moreover, the UD model selection can be faster by parallelization using different sets of parameters. The multilevel SVM-based algorithms can be applied on real-world applications. 

% A parallel implementation of SVMs with bagging method can be used to accelerate the model training process on large datasets. 

\bibliographystyle{plain} 
\bibliography{paper,ilya}

\end{document}